\documentclass{article}
\usepackage{graphicx} 
\usepackage{amsmath}
\usepackage{amssymb}
\usepackage{booktabs}
\usepackage{placeins}
\usepackage{multirow}
\title{URS-Stereo: Uncertainty-Guided Residual Search for Real-Time Stereo
Matching}
\author{Pouya Sohrabipour, 
Chaitanya kumar reddy Pallerla, Dongyi Wang}

\usepackage{comment}
\usepackage{xcolor}
\begin{document}

\maketitle


\begin{abstract}
Real-time stereo matching is crucial for robotics, autonomous systems, and embedded vision applications, where both computational efficiency and disparity accuracy are required. Recent coarse-to-fine stereo matching methods improve efficiency by progressively refining disparity estimates using local cost volumes at higher resolutions. However, these methods rely heavily on the accuracy of propagated disparity estimates from previous stages. When the propagated disparity is inaccurate, the ground-truth correspondence may fall outside the predefined local search range, leading to unrecoverable matching failures during subsequent refinement.

In this paper, we propose URS-Stereo, a real-time coarse-to-fine stereo matching framework that addresses this limitation through uncertainty-guided search adaptation. Specifically, we introduce an Uncertainty-Guided Residual Search Module (UGRSM), which predicts the reliability of propagated disparities together with residual search offsets to adaptively relocate the centers of local cost volumes before disparity refinement. By dynamically adjusting the search region according to the confidence of the propagated disparity, the proposed method significantly improves the robustness of local correspondence estimation while preserving the computational efficiency of coarse-to-fine stereo matching.

Extensive experiments on SceneFlow, KITTI 2012, KITTI 2015, Middlebury, and ETH3D demonstrate that URS-Stereo consistently improves disparity estimation while maintaining real-time inference speed, validating the effectiveness of the proposed uncertainty-guided search strategy.
\end{abstract}

\section{Introduction}
\label{sec:intro}


Stereo matching remains one of the most effective approaches for dense depth estimation due to its strong geometric foundation. Recent deep learning methods construct cost volumes and learn dense disparity estimation in an end-to-end manner, achieving remarkable accuracy. However, constructing dense cost volumes at high image resolutions over large disparity ranges incurs substantial computational and memory costs, making many high-performance stereo networks unsuitable for real-time applications~\cite{xu2023iterative,wen2025foundationstereo,liu2024goat,wei2025waveletstereo,wang2025promptstereo,chen2024mochastereo}.

To improve efficiency, modern stereo matching methods commonly adopt a coarse-to-fine strategy, where disparity is first estimated at a coarse resolution and progressively refined at higher resolutions using local cost volumes centered around the propagated disparity estimate~\cite{cheng2024monsterpp,lipson2021raft,wang2019anynet}. Compared with constructing a full cost volume at every scale, this significantly reduces both computation and memory by limiting the disparity search range during refinement.

Despite its efficiency, coarse-to-fine stereo matching suffers from an inherent limitation. The local cost volume at each refinement stage is constructed around the upsampled disparity predicted by the previous stage. Consequently, the quality of disparity refinement depends entirely on the accuracy of this propagated disparity. If the coarse prediction or its upsampling introduces a sufficiently large error, the ground-truth disparity may fall outside the predefined local search range. Once this occurs, the correct correspondence is no longer contained in the cost volume, preventing subsequent refinement stages from recovering the true disparity regardless of the effectiveness of the matching network. This issue is particularly severe around object boundaries, thin structures, occluded regions, and areas with large disparity variations, where disparity propagation errors are more likely to occur.

To address this limitation, we propose URS-Stereo, a real-time coarse-to-fine stereo matching framework that introduces an \emph{Uncertainty-Guided Residual Search Module (UGRSM)}. Instead of assuming that the propagated disparity always provides an accurate search center, UGRSM estimates the reliability of the propagated disparity and predicts an uncertainty-guided residual offset to adaptively adjust the center of the local cost volume before disparity refinement. By dynamically relocating the search window according to the confidence of the propagated estimate, URS-Stereo substantially reduces the probability that the ground-truth disparity falls outside the local search range, leading to more robust disparity refinement while preserving the computational efficiency of coarse-to-fine stereo matching.

The main contributions of this work are summarized as follows:

\begin{itemize}
\item We analyze the fundamental limitation of coarse-to-fine stereo matching and show that inaccurate disparity propagation can shift the local cost volume away from the ground-truth correspondence, resulting in unrecoverable matching failures.

\item We propose the Uncertainty-Guided Residual Search Module (UGRSM), which predicts uncertainty-aware residual search offsets to adaptively relocate local cost-volume centers based on the reliability of propagated disparities.

\item Extensive experiments on SceneFlow, KITTI 2012, KITTI 2015, Middlebury, and ETH3D demonstrate that URS-Stereo consistently improves disparity estimation while maintaining real-time inference speed.
\end{itemize}

\begin{figure*}[t]
\centering
\includegraphics[width=\textwidth]{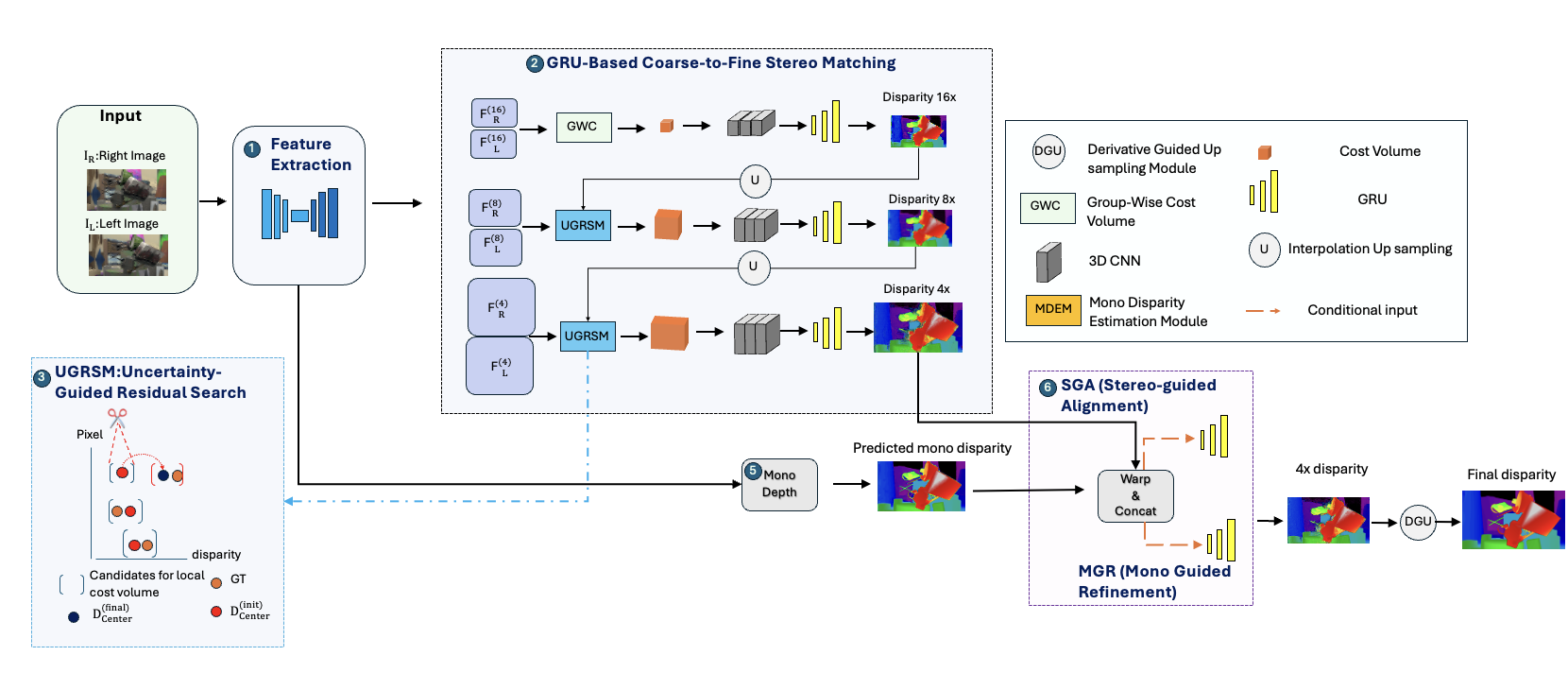}
\caption{
Overview of the proposed URS-Stereo framework. A full cost volume is constructed at the $\frac{1}{16}$ resolution, while uncertainty-guided local cost volumes are constructed at the $\frac{1}{8}$ and $\frac{1}{4}$ resolutions. The proposed UGRSM predicts uncertainty-aware residual search offsets to adjust local search centers. MDEM provides lightweight monocular disparity guidance through a teacher-student framework. Finally, DEM reconstructs the full-resolution disparity map using learned disparity derivatives.
}
\label{fig:main}
\end{figure*}
\section{RELATED WORKS }

Deep depth estimation methods can be broadly categorized into three groups: optical triangulation/stereo matching methods, monocular depth estimation methods, and hybrid mono--stereo approaches. Each category addresses depth estimation from a different perspective, with distinct trade-offs between accuracy, efficiency, and robustness.

\subsection{Stereo Matching Based Depth Estimation}

Stereo matching estimates depth by exploiting geometric correspondences between rectified image pairs. Early stage deep learning approaches replaced handcrafted matching costs with learned representations, demonstrating the effectiveness of convolutional neural networks for disparity estimation 
\cite{kendall2017end,zhang2019ganet,shen2021cfnet,mao2021uasnet,zhao2023high,feng2023mc,xu2023unifying,zbontar2015mccnn,mayer2016dispnet}.
These methods construct a cost volume over disparity candidates and regress dense disparity maps in an end-to-end manner. 

Subsequent works focused on improving accuracy through cost-volume regularization using 3D convolutional neural networks \cite{guo2019gwcnet,bangunharcana2021correlate,cheng2020hierarchical,Duggal2019ICCV,wen2026fastfoundationstereo}. Pyramid Stereo Matching Network (PSMNet) \cite{Chang_Che} introduced spatial pyramid pooling to enlarge the receptive field, while later methods such as GA-Net \cite{zhang2019ganet} and cascade-based architectures \cite{Gu_Fan_Zhu_Dai_Tan_2020_CascadeCostVolume} further enhanced geometric consistency at the cost of increased memory and computational demands.

To alleviate the heavy computational cost of 3D convolutions, recent methods adopt iterative refinement strategies. RAFT-Stereo \cite{lipson2021raft} and iterative geometry encoding approaches \cite{xu2023iterative} generate an initial disparity estimate and refine it using recurrent update operators, significantly reducing memory usage while maintaining competitive accuracy. However, these methods still face challenges in achieving real-time performance under strict resource constraints.

Recent efforts have specifically focused on real-time stereo matching \cite{Yan2025MatchAttention,Xu2026MAFNet,Jing_2026_CVPR}. Cheng et al.~\cite{Cheng_Yang_Li_2024_XR_Stereo} proposed Stereo Matching in Time, which employs a recurrent update mechanism to achieve extremely high inference speeds. While highly efficient, such recurrent approaches often face challenges in maintaining disparity accuracy and generalization under strict computational constraints. Another representative real-time method is HITNet~\cite{tankovich2021hitnet}, which adopts a hierarchical architecture that progressively refines disparity estimates across multiple scales using confidence-guided updates. Although effective, the construction and maintenance of multi-scale cost representations can lead to increased memory consumption, limiting deployment on lightweight platforms.

\subsection{Monocular Depth Estimation}

Monocular depth estimation predicts scene depth from a single RGB image by learning semantic and contextual cues such as perspective, object size, and texture gradients \cite{Godard2017Monodepth,monodepth2,Ranftl2022,Birkl2023MiDaSv31,Yang2024DepthAnything}. Early deep learning models employed multi-scale convolutional architectures to regress dense depth maps \cite{eigen2014depth,laina2016deeper}. Unsupervised and self-supervised approaches later leveraged stereo consistency during training to reduce reliance on ground-truth depth annotations \cite{godard2017unsupervised}.

More recently, large-scale pretraining and vision foundation models have significantly advanced monocular depth estimation. Depth Anything V2 \cite{depth_anything_v2} demonstrates strong generalization across diverse scenes, while self-supervised representation learning methods such as DINOv2 \cite{oquab2023dinov2} provide robust visual features that benefit dense prediction tasks. Despite their strong semantic understanding, monocular methods lack metric depth accuracy and often struggle with scale ambiguity.\cite{Bhat2023ZoeDepth,tri-zerodepth,Zeng2024RSA}

\subsection{Hybrid Mono--Stereo Methods}

Hybrid approaches aim to combine the geometric precision of stereo matching with the semantic robustness of monocular depth estimation. MonSter \cite{cheng2025monster} integrates monocular and stereo depth predictions through dual refinement branches, identifying unreliable regions using flaw maps and selectively correcting them via cross-modal guidance. While highly accurate, its multi-stage architecture introduces substantial computational overhead.

MonSter \cite{cheng2025monster} integrates monocular and stereo depth predictions through dual refinement branches. Specifically, it generates disparity maps from both modalities and employs flaw maps to identify unreliable regions. Stereo-guided and mono-guided refinement branches are then used to correct these regions through cross-modal interaction. While highly accurate, its multi-stage architecture introduces substantial computational overhead and limits real-time applicability.

To improve efficiency, RT-MonSter \cite{cheng2024monsterpp} extends this hybrid framework using a coarse-to-fine design. A dense cost volume is constructed only at the coarsest resolution, while local cost volumes at finer scales are centered around the propagated disparity estimate from the previous stage. This strategy significantly reduces the disparity search space and improves inference speed. However, its effectiveness depends heavily on the accuracy of disparity propagation, since errors introduced at coarse resolutions may lead to incorrect local search centers at finer scales.

These limitations motivate the development of more efficient hybrid frameworks that preserve geometric fidelity during upsampling while adaptively leveraging monocular cues only when beneficial.

\section{Our Approach}

\subsection{Overview}

We propose URS-Stereo, a real-time coarse-to-fine stereo matching framework designed to improve the robustness of disparity refinement. The overall architecture of URS-Stereo is illustrated in Fig.~\ref{fig:main}.

The key component of URS-Stereo is the proposed Uncertainty-Guided Residual Search Module (UGRSM). Instead of directly constructing local cost volumes around the propagated disparity estimate, UGRSM predicts the reliability of the propagated disparity together with a residual search offset to adaptively adjust the center of the local cost volume. This uncertainty-guided search strategy reduces the risk that the ground-truth disparity falls outside the local search range due to propagation errors, enabling more robust disparity refinement while preserving the computational efficiency of coarse-to-fine stereo matching.

\subsection{Multi-Scale Feature Extraction}

To start, given the input stereo pair $I_L$ and $I_R$, a shared CNN-based feature extractor produces multi-scale feature maps for both images. We denote the extracted left and right features at scale $s$ as $F_L^{(s)}$ and $F_R^{(s)}$, where $s \in {16,8,4}$ represents the downsampling factor relative to the input image resolution. These features are used for cost-volume construction, GRU-based coarse-to-fine stereo matching, uncertainty-guided residual Search, residual and mono-disparity prediction. 

\subsection{GRU-Based Coarse-to-Fine Stereo Matching}

Coarse-to-fine stereo matching has become a widely adopted strategy for real-time disparity estimation because it significantly reduces the computational cost of cost-volume construction by progressively refining disparity predictions across multiple resolutions~\cite{cheng2024monsterpp,lipson2021raft,xu2023iterative}. Following this paradigm, URS-Stereo estimates disparity in a coarse-to-fine manner using three stages operating at the $\frac{1}{16}$, $\frac{1}{8}$, and $\frac{1}{4}$ image resolutions. Disparity is first estimated at a coarse scale and subsequently refined at finer scales using progressively smaller search ranges.

At the coarsest $\frac{1}{16}$ scale, a full group-wise correlation (GWC) cost volume is constructed between the left and right matching features. Let $F_L^{(16)}$ and $F_R^{(16)}$ denote the left and right feature maps, respectively. The channels are divided into $G$ groups, and the group-wise correlation volume is computed as

\begin{equation}
C_g(d,x,y)
=
\frac{1}{N_g}
\left\langle
F_{L,g}^{(16)}(x,y),
F_{R,g}^{(16)}(x-d,y)
\right\rangle
\end{equation}

where $g$ denotes the group index, $N_g$ is the number of channels in each group, and $\langle \cdot,\cdot \rangle$ denotes the inner product between feature vectors.

The resulting cost volume is regularized using lightweight 3D convolutional layers and converted into a disparity probability distribution through a softmax operation:

\begin{equation}
P(d,x,y)
=
\frac{\exp(-C_g(d,x,y)}
{\sum_{d'} \exp(-C_g(d',x,y))}
\end{equation}

The initial disparity estimate, denoted by $D_0^{(16)}$, represents the first disparity prediction at the $\frac{1}{16}$ scale. It is obtained using disparity regression:

\begin{equation}
D_0^{(16)}(x,y)
=
\sum_{d=0}^{D_{\max}/16}
d \cdot P(d,x,y)
\end{equation}

The disparity obtained through regression provides an initial estimate but may still contain matching ambiguities and local errors. Therefore, following recent iterative stereo matching frameworks~\cite{lipson2021raft,xu2023iterative}, URS-Stereo employs a ConvGRU update block to progressively refine the disparity estimate at the $\frac{1}{16}$ resolution.

Following RAFT-Stereo~\cite{lipson2021raft} and IGEV~\cite{xu2023iterative}, disparity refinement is formulated as a recurrent optimization process using ConvGRUs. Let $H_i^{(s)}$ denote the hidden state at iteration $i$ and scale $s$, and let $G_i^{(s)}$ denote the geometry feature retrieved from the encoded cost volume. The ConvGRU updates its hidden state according to
\begin{equation}
z_i
=
\sigma
\left(
\mathrm{Conv}
\left(
[H_{i-1}^{(s)},G_i^{(s)}],W_z
\right)+c_k
\right)
\end{equation}

\begin{equation}
r_i
=
\sigma
\left(
\mathrm{Conv}
\left(
[H_{i-1}^{(s)},G_i^{(s)}],W_r
\right)+c_r
\right)
\end{equation}

\begin{equation}
\tilde{H}_i^{(s)}
=
\tanh
\left(
\mathrm{Conv}
\left(
[r_i \odot H_{i-1}^{(s)},G_i^{(s)}],W_h
\right)+c_h
\right)
\end{equation}

\begin{equation}
H_i^{(s)}
=
(1-z_i)\odot H_{i-1}^{(s)}
+
z_i\odot \tilde{H}_i^{(s)}
\end{equation}

where $c_k$ , $c_r$ , $c_h$ are context features, $H_i^{(s)}$ denotes the ConvGRU hidden state and $\odot$ denotes element-wise multiplication. Based on the updated hidden state, a disparity decoder predicts a residual disparity update $\Delta D_i^{(s)}$, which is used to refine the current disparity estimate:

\begin{equation}
D_{i+1}^{(s)}
=
D_i^{(s)}
+
\Delta D_i^{(s)}
\end{equation}

After $N_s$ recurrent update iterations, the final disparity prediction at scale $s$ is denoted by $D^{(s)}$.

The refined disparity is then propagated to the $\frac{1}{8}$ and $\frac{1}{4}$ stages.

At finer scales, constructing a full cost volume becomes computationally expensive because the cost volume scales with both image resolution and disparity search range. As the spatial resolution increases, maintaining a dense cost volume over all disparity candidates results in substantial memory and computational overhead. Therefore, URS-Stereo employs a newly proposed Uncertainty-Guided Residual Search Module (UGRSM) to predict an adaptive search center and construct a local cost volume around the propagated disparity estimate. The resulting local cost volume is encoded by the geometry encoding module and used by the ConvGRU update block for disparity refinement. This coarse-to-fine strategy enables efficient disparity estimation while maintaining robustness to disparity propagation errors.

\subsection{Uncertainty-Guided Residual Search Module}

The UGRSM module is designed to reduce local search failure in coarse-to-fine stereo matching. At finer scales, local cost volumes are built around a predicted disparity center. If this center is inaccurate, the ground-truth disparity may fall outside the local search window.

\begin{figure}[t]
\centering
\includegraphics[width=\linewidth]{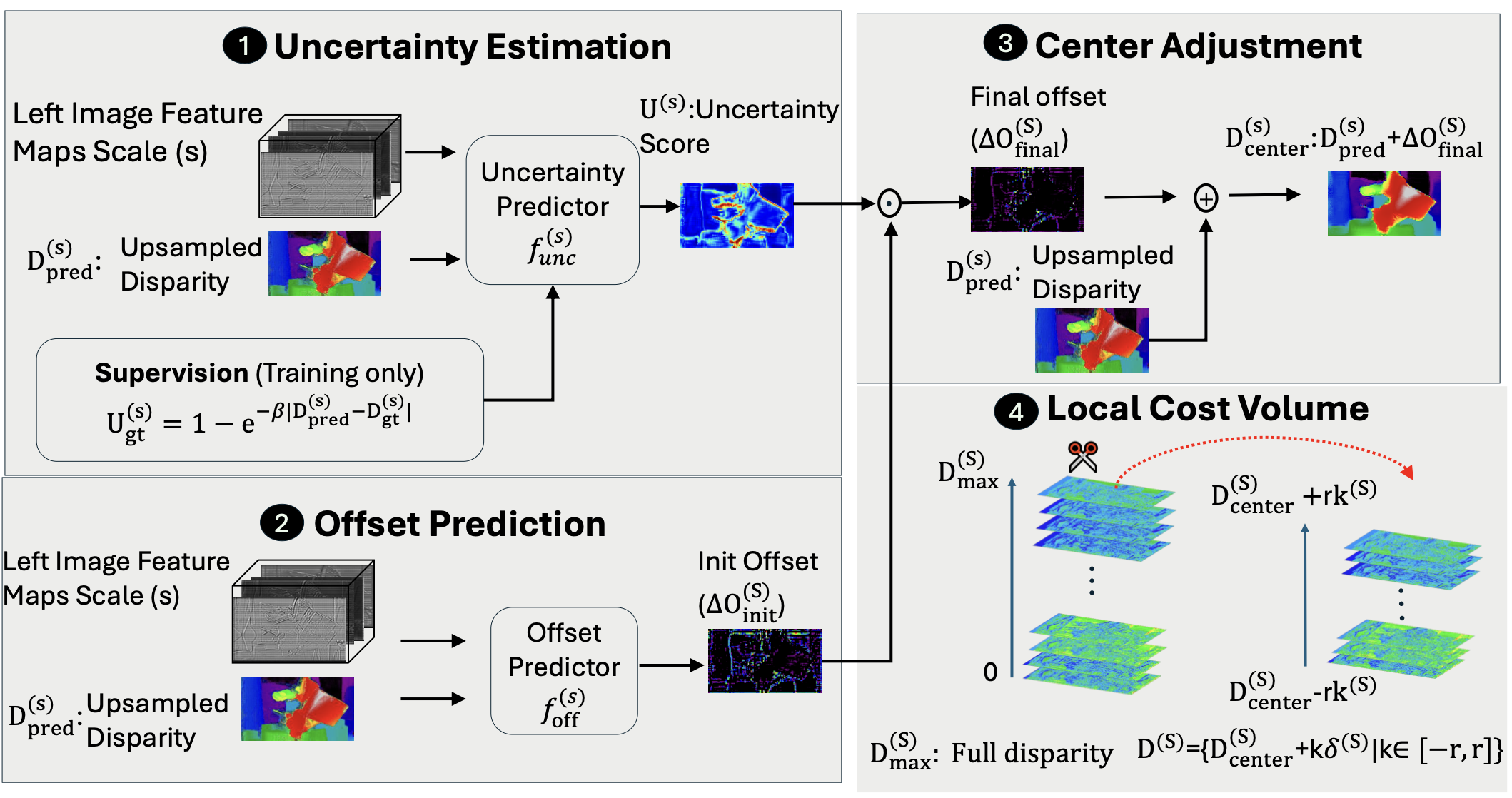}
\caption{
Architecture of the proposed Uncertainty-Guided Residual Search Module (UGRSM). The module predicts an uncertainty map and a residual search offset from the propagated disparity and image features. The uncertainty score modulates the residual offset to produce an adjusted local cost-volume center used for local disparity search.
}
\label{fig:ugrsm}
\end{figure}

To estimate the reliability of the propagated disparity, UGRSM first predicts an uncertainty map. Given the left image features $F_L^{(s)}$ and the upsampled disparity estimate $D^{(s)}$ propagated from the previous scale, the uncertainty predictor estimates

\begin{equation}
U^{(s)}
=
f_{unc}^{(s)}
\left(
F_L^{(s)},
D^{(s)}
\right)
\end{equation}

where $U^{(s)} \in [0,1]$.

To training $f_{unc}^{(s)}$ in a end-to-end maner, considering ground-truth uncertainty labels are unavailable, pseudo uncertainty targets are generated from disparity prediction error:

\begin{equation}
U_{gt}^{(s)}
=
1
-
\exp
\left(
-\beta
\left|
D_{pred}^{(s)}
-
D_{gt}^{(s)}
\right|
\right)
\end{equation}

where $D_{pred}^{(s)}$ denotes the upsampled disparity propagated from the previous scale $\frac{s}{2}$, $D_{gt}^{(s)}$ denotes the ground-truth disparity at scale $s$, and hyperparameter $\beta$ controls the sensitivity of the uncertainty target to disparity error.

With $U^(s)$ predicted, let $O^{(s)}$ denote the center of the local cost volume at scale $s$. The goal of UGRSM is to estimate a residual offset $\Delta O^{(s)}$ that adjusts this center when the propagated disparity is unreliable.

In the second step of UGRSM, an initial residual search offset, denoted by $\Delta O_{init}^{(s)}$, will be predicted using a lightweight pixel-wise network, which represents the displacement of the local cost-volume center from the propagated disparity estimate. The network can be expressed as:
\begin{equation}
\Delta O_{init}^{(s)}
=
f_{off}^{(s)}
\left(
F_L^{(s)},
D^{(s)}
\right)
\end{equation}

In the step three, the final offset prediction will modulate the initial offset by the predicted uncertainty from step 1:

\begin{equation}
\Delta O_{final}^{(s)}
=
U^{(s)}
\cdot
\Delta O_{init}^{(s)}
\end{equation}
The adjusted search center is then defined as

\begin{equation}
D_{center}^{(s)}
=
D_{pred}^{(s)}
+
\Delta O_{final}^{(s)}
\end{equation}

In the step four, The adjusted search center is used to generate a set of local disparity candidates:

\begin{equation}
\mathcal{D}^{(s)}
=
\left\{
D_{center}^{(s)}
+
k\delta^{(s)}
\;\middle|\;
k \in [-r,r]
\right\}
\end{equation}

where $r$ denotes the search radius and $\delta^{(s)}$ is the disparity interval at scale $s$.

In URS-Stereo, UGRSM is applied at the $\frac{1}{8}$ and $\frac{1}{4}$ stages, where local cost volumes are used. At the $\frac{1}{16}$ stage, a full cost volume is constructed, so residual search offset prediction is not required. This design allows confident pixels to preserve their original search center, while uncertain pixels are allowed to shift the search window toward a more reliable disparity range.
\subsection{Mono-Stereo Mutual Refinement}

At the final $\frac{1}{4}$ stage, URS-Stereo performs mono-stereo mutual refinement. The stereo disparity and mono disparity are used to warp the right image features toward the left image. The corresponding feature differences define stereo and mono flaw cues:

\begin{equation}
E_{stereo}
=
F_L^{(4)}
-
\mathcal{W}
\left(
F_R^{(4)},
D_{stereo}^{(4)}
\right)
\end{equation}

\begin{equation}
E_{mono}
=
F_L^{(4)}
-
\mathcal{W}
\left(
F_R^{(4)},
D_{mono}^{(4)}
\right)
\end{equation}

where $\mathcal{W}(\cdot)$ denotes disparity-based warping.

The stereo update branch uses stereo disparity, mono disparity, geometric features, and flaw cues to refine the stereo disparity. Similarly, the mono update branch refines the mono disparity using the stereo branch as guidance. This mutual refinement allows the model to combine reliable stereo matching cues with complementary mono-disparity information.

\subsection{Training Objective}

The proposed network is trained end-to-end using stereo disparity supervision and uncertainty supervision.

The stereo matching branch is supervised at every recurrent iteration using an exponentially weighted disparity loss:

\begin{equation}
\mathcal{L}_{stereo}
=
\sum_{i=1}^{N}
\gamma^{N-i}
\|D_i-D_{gt}\|_1
\end{equation}

where $D_i$ denotes the disparity prediction at iteration $i$, $D_{gt}$ is the ground-truth disparity, and $\gamma$ is a decay factor that assigns larger weights to later predictions.

The uncertainty prediction branch is supervised using the pseudo uncertainty target:

\begin{equation}
\mathcal{L}_{unc}
=
\sum_{i=1}^{N_u}
\gamma^{N_u-i}
\|U_i-U_{gt}\|_1
\end{equation}

where $U_i$ and $U_{gt}$ denote the predicted and target uncertainty maps, respectively.

The overall training objective is defined as

\begin{equation}
\mathcal{L}
=
\mathcal{L}_{stereo}
+
\lambda_{unc}\mathcal{L}_{unc},
\end{equation}

where $\lambda_{unc}$ controls the contribution of the uncertainty supervision.

\section{Experiment}

To evaluate the generalization capability of our method, we conduct experiments on several widely used stereo matching benchmarks, including KITTI 2012, KITTI 2015 and Middlebury. URS-Stereo is trained exclusively on the SceneFlow dataset and directly evaluated on all real-world benchmarks without any dataset-specific fine-tuning. We compare our method against recent efficient stereo matching approaches under the same SceneFlow-only training setting.

\textbf{KITTI 2012 and KITTI 2015.}
We evaluate the zero-shot generalization performance of URS-Stereo on the KITTI 2012 \cite{geiger2012kitti} and KITTI 2015 \cite{menze2015object} benchmarks. The model trained exclusively on SceneFlow is directly evaluated on both KITTI datasets without any additional fine-tuning. These benchmarks contain real-world driving scenes with diverse illumination conditions, reflective surfaces, and challenging outdoor environments. Quantitative results on both KITTI benchmarks are reported in Tab.~\ref{tab:benchmark}.

\textbf{Middlebury.}
We additionally evaluate URS-Stereo on the Middlebury ~\cite{scharstein2002taxonomy} benchmark, which contains high-resolution indoor stereo pairs with challenging lighting conditions, fine structures, and large textureless regions. The model trained exclusively on SceneFlow is directly evaluated without dataset-specific fine-tuning. Quantitative results are presented in Tab.~\ref{tab:benchmark}.

\begin{table}[t]
\centering

\caption{Zero-shot generalization results on KITTI 2012, KITTI 2015, and Middlebury. Best results are shown in \textcolor{red}{red}, and second-best results are shown in \textcolor{orange}{orange}.}
\label{tab:benchmark}
\resizebox{\linewidth}{!}{
\begin{tabular}{lcccccc}
\toprule
\multirow{2}{*}{Method} 
& \multicolumn{2}{c}{KITTI 2012} 
& \multicolumn{2}{c}{KITTI 2015} 
& \multicolumn{2}{c}{Middlebury} \\
\cmidrule(lr){2-3}
\cmidrule(lr){4-5}
\cmidrule(lr){6-7}
& D1 & EPE & D1 & EPE & Bad 2.0 & EPE \\
\midrule
\multicolumn{7}{l}{\textit{Efficient methods: SceneFlow}} \\
\midrule
CoEX~\cite{bangunharcana2021correlate}                & 22.30 & 3.37 & 17.33 & 3.00 & 26.42 & 4.90 \\
MobileStereoNet-2D~ \cite{shamsafar2022mobilestereonet} & 19.30 & 2.51 & 21.88 & 2.85 & 37.98 & 7.54 \\
MobileStereoNet-3D~ \cite{shamsafar2022mobilestereonet}& 19.59 & 2.67 & 17.89 & 3.01 & 25.26 & 4.61 \\
FastACV~ \cite{xu2022attention}         & 13.90 & 2.05 & 11.83 & 2.21 & 19.61 & 4.66 \\
FastACV+~  \cite{xu2024accurate}    & 16.69 & 2.29 & 15.28 & 3.27 & 27.34 & 7.16 \\
Lite-CREStereo++~ \cite{jing2023uncertainty} & 5.93 & 1.29 & 7.37 & 1.36 & 14.91 & 3.32 \\
LightStereo-M~ \cite{guo2025lightstereo} & 6.76 & 1.26 & 6.79 & 1.42 & 16.99 & 2.06 \\
LightStereo-L \cite{guo2025lightstereo} ~& 6.80 & 1.27 & 6.62 & 1.29 & 17.23 & 2.88 \\
BANet-2D~  \cite{xu2025banet}         & 14.75 & 2.23 & 16.98 & 3.91 & 26.79 & 6.96 \\
BANet-3D~   \cite{xu2025banet}        & 17.39 & 2.41 & 17.28 & 3.36 & 28.79 & 8.05 \\
Lite Any Stereo \cite{jing2026lite}              & 5.45 & 1.18 & 6.45 & 1.32 & 13.13 &\textcolor{red} {1.60} \\
\midrule
\textbf{ours }                        & \textcolor{red}{3.98} & \textcolor{red}{0.89} & \textcolor{red}{5.85} &\textcolor{red}{1.24}  & \textcolor{red}{13.03} & \textcolor{orange}{1.71 }\\
\bottomrule
\end{tabular}
}
\end{table}

The offset-only variant improves disparity estimation by allowing the local search center to deviate from the propagated disparity estimate. However, applying residual offsets without explicitly considering the reliability of the propagated disparity may unnecessarily modify search centers in already reliable regions. By incorporating uncertainty guidance, UGRSM adaptively modulates the predicted residual offset according to the estimated reliability of the propagated disparity. This results in more effective search-center adjustment and further improves disparity accuracy with minimal additional computational overhead.

\FloatBarrier
\section{Conclusion}

In this paper, we presented URS-Stereo, a real-time coarse-to-fine stereo matching framework designed to improve the robustness of local disparity refinement. We identified a fundamental limitation of coarse-to-fine stereo matching: inaccurate disparity propagation can shift the center of a local cost volume away from the ground-truth correspondence, preventing subsequent refinement stages from recovering the correct disparity. To address this issue, we introduced the Uncertainty-Guided Residual Search Module (UGRSM), which estimates the reliability of propagated disparities and predicts uncertainty-guided residual offsets to adaptively relocate local cost-volume centers. This enables the search region to dynamically adjust to propagation errors while preserving the computational efficiency of local cost-volume construction.

Extensive experiments on SceneFlow, KITTI 2012, KITTI 2015, Middlebury, and ETH3D demonstrate that URS-Stereo achieves robust disparity estimation and strong zero-shot generalization while maintaining real-time inference speed. These results validate the effectiveness of uncertainty-guided search adaptation for improving coarse-to-fine stereo matching. We believe that adaptively modeling the reliability of propagated predictions provides a promising direction for developing efficient and robust stereo matching systems.

\bibliographystyle{ieeetr}
\bibliography{main}

\end{document}